\documentclass[runningheads]{llncs}
\usepackage[T1]{fontenc}
\usepackage{graphicx}

\usepackage[misc]{ifsym}
\usepackage{amsmath}
\usepackage{amssymb}
\usepackage{hyperref}
\usepackage{booktabs}
\usepackage{algorithm}
\usepackage{algpseudocode}
\usepackage{multirow}
\usepackage{tabularx}
\usepackage{bm}
\usepackage{caption}
\usepackage{wrapfig}
\usepackage{float}

\usepackage{orcidlink}

\begin{document}
\pagenumbering{gobble}

\title{NodeNAS: Node-Specific Graph Neural Architecture Search for Out-of-Distribution Generalization}
\titlerunning{Node-Specific Graph Neural Architecture Search}
%
\hypersetup{hidelinks,
	colorlinks=true,
	allcolors=black,
	pdfstartview=Fit,
	breaklinks=true}

\author{Qiyi Wang\inst{\ast}~\orcidlink{0009-0001-4605-6830} \and
Yinning Shao\inst{\ast}~\orcidlink{0009-0000-6555-0802} \and
Yunlong Ma\inst{\dagger} \textsuperscript{(\Letter)} ~\orcidlink{0000-0001-9947-7746} \and
Min Liu\inst{}~\orcidlink{0000-0002-8902-5460}}


\authorrunning{Q. Wang et al.}
\institute{Tongji University, Shanghai 201800, China
\email{\{wqy126179,yinningshao,evanma,lmin\}@tongji.edu.cn}
}
%
\maketitle              
\begin{abstract}
Graph neural architecture search (GraphNAS) has demonstrated advantages in mitigating performance degradation of graph neural networks (GNNs) due to distribution shifts. Recent approaches introduce weight sharing across tailored architectures, generating unique GNN architectures for each graph end-to-end. 
However, existing GraphNAS methods do not account for distribution patterns across different graphs and heavily rely on extensive training data.
With sparse or single training graphs, these methods struggle to discover optimal mappings between graphs and architectures, failing to generalize to out-of-distribution (OOD) data.
In this paper, we propose node-specific graph neural architecture search (NodeNAS), which aims to tailor distinct aggregation methods for different nodes by disentangling node topology and graph distribution with limited datasets.
We further propose adaptive aggregation attention-based Multi-dim NodeNAS method (MNNAS), which learns a node-specific architecture customizer with good generalizability.
Specifically, we extend the vertical depth of the search space, supporting simultaneous customization of the node-specific architecture across multiple dimensions.
Moreover, we model the power-law distribution of node degrees under varying assortativity, encoding structure-invariant information to guide architecture customization across each dimension.
Extensive experiments across supervised and unsupervised tasks demonstrate that MNNAS surpasses state-of-the-art algorithms and achieves excellent OOD generalization.

\end{abstract}
\footnotetext[1]{* Both authors contributed equally to this paper.}
\footnotetext[2]{† Corresponding author.}

\section{Introduction}
Graph Neural Networks (GNNs) demonstrate exceptional performance in a variety of graph-based tasks\cite{xu2018powerful,nazi1903gap,wilder2019end} including graph classification, graph partitioning, and community detection.
However, the performance of GNNs is based on the message passing mechanism, which operates by aggregating information from the local neighborhood of nodes.
Consequently, the performance of trained GNNs heavily relies on the local structural characteristics of the graph, creating a dependency that predisposes these models to overfitting and significantly undermines performance when faced with distribution shifts.

Graph Neural Architecture Search (GraphNAS) has recently demonstrated significant potential to mitigate performance degradation caused by distribution shifts.
By automating the architectural engineering process and exploring a wide range of candidate solutions, GraphNAS facilitates the autonomous discovery of optimal designs for GNNs. 
The early GraphNAS framework \cite{gao2021graph,zhou2022auto} formulates the search for optimal architectures as a black-box optimization problem over a discrete search space, which inherently restricts its effectiveness to scenarios with independent and identically distributed (IID) data distributions.
The latest work generates distinct architectures for each graph based on learned graph representations through differentiable search and weight sharing\cite{qin2022graph,yao2024data}.
They relax the rigid selection of the candidate set to a weighted combination of all candidates and share the weights of candidate operations, resulting in outstanding performance on out-of-distribution (OOD) data.

However, GraphNAS still faces the following limitations: 
{(1)}
Existing GraphNAS methods heavily rely on large amounts of training data to facilitate the model to capture the preferences of different graph instances for GNN architectures. Current GraphNAS methods struggle to discover the optimal mapping between graphs and architectures when the training graphs are sparse or even singular. 
{(2)} 
Existing GraphNAS methods generally tailor graph-specific architectures that use the same aggregation method for all nodes in the graph.
However, a practical barrier to the generalization of tailored architectures arises from the long-tail node degree distribution present in many large-scale real-world graphs.
Currently no GraphNAS method offers distinct aggregation strategies tailored for high-degree (head) and low-degree (tail) nodes.


\begin{figure}[t]
    \centering
    \includegraphics[width=0.75\linewidth]{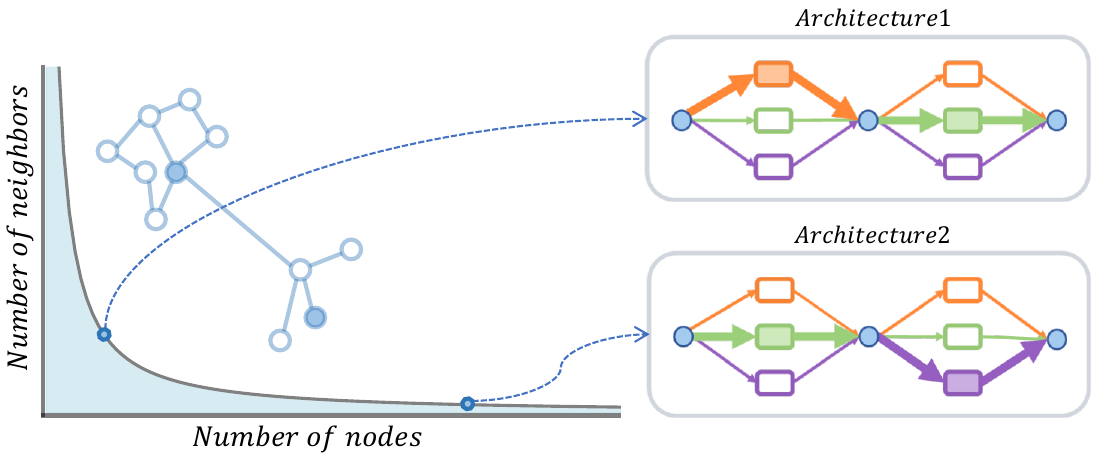}
    \caption{An overview of NodeNAS}
    \label{fig:intro_draw}
\end{figure}

To address these challenges, we propose Node-specific graph Neural Architecture Search (NodeNAS), aimed at end-to-end mapping nodes in long-tail degree distribution to specific architectures. 
Specifically, NodeNAS searches for a unique probability vector for each node, where each value in the vector represents the probability of a candidate operation.
NodeNAS is designed to identify optimal embedding-update methods for different nodes under graph distributions with varying assortativity, facilitating model generalization to OOD graphs.
This approach necessitates disantangling degree distribution from graph type in the learning process of probability vectors to counteract spurious motifs.

Further, we propose Multi-dimension NodeNAS method (MNNAS), which captures structure-invariant factors hidden within the graph and tailors node-specific architectures for graphs with unknown distributions. 
Specifically, we first introduce a mapping encoder that maps different operations to distinct embeddings and projects them into the architecture search space. 
We propose a multi-dimension architecture search network with differentiable operation mixture weights, extending the search space through multiple Search Dimensions (S-Dims) to enable cross-dimensional optimization.
To reduce the scale of learnable parameters, different S-Dims for each node are designed to share the same operation-embedding mapping. 
Meanwhile, we design adaptive aggregation attention with a link pattern encoder to capture distribution commonalities across graph assortativities, while identifying spurious motifs during architecture search.
Guided by the encoder, the attention mechanism customizes multiple node-specific architectures across various graph topologies in parallel, avoiding performance limitations from single-dimension strategies.
Finally, architectures tailored from multiple dimensions are integrated to generate node representations with generalization. 
By sharing weights across different architectures, MNNAS can tailor multi-dim node-specific architectures end-to-end and output results for downstream tasks. 
Using information bottleneck (IB) theory, we demonstrate the interpretability of MNNAS in OOD generalization.
Extensive experiments on unsupervised and supervised tasks further validate the superiority of MNNAS over benchmark methods.

Our contributions can be summarized as follows.
\begin{itemize}
    \item 
    We propose a novel node-specific graph neural architecture search method that tailors embedding update strategies for nodes, enabling flexible adaptation to graphs with unknown distributions.
    \item 
    We design the adaptive aggregation attention that disentangles power-law degree distribution from distinct assortativity and propose multi-dim architecture search network for architecture customization. MNNAS could interpretably tailors high-performing node-specific architectures for OOD graphs.
    \item 
    To the best of our knowledge, MNNAS is the first NAS model to exhibit OOD generalization even with single-graph training, and the first NAS model to be applied to unsupervised tasks such as community detection while demonstrating good OOD generalization.
\end{itemize}

\section{Preliminaries}
\subsection{Out-Of-Distribution Generalization}
Given the graph space $\mathcal{G}$ and label space $\mathcal{Y}$, we define a training graph dataset $\mathcal{G}_{\text{tr}}=\{g_i\}_{i=1}^{N_{\text{tr}}}, g_i \in \mathcal{G}$, along with a corresponding label set $\mathcal{Y}_{\text{tr}}=\{y_i\}_{i=1}^{N_{\text{tr}}}, y_i \in \mathcal{Y}$. Similarly, the test graph dataset is denoted as $\mathcal{G}_{\text{te}}=\{g_i\}_{i=1}^{N_{\text{te}}}$, and the label set as $\mathcal{Y}_{\text{te}}=\{y_i\}_{i=1}^{N_{\text{te}}}$. 
The objective of out-of-distribution generalization is to achieve a model $F: \mathcal{G} \to \mathcal{Y}$ using $\mathcal{G}_{\text{tr}}$ and $\mathcal{Y}_{\text{tr}}$, which performs effectively on $\mathcal{G}_{\text{te}}$ and $\mathcal{Y}_{\text{te}}$, under the assumption that the distributions $P(\mathcal{G}_{\text{tr}}, \mathcal{Y}_{\text{tr}}) \neq P(\mathcal{G}_{\text{te}}, \mathcal{Y}_{\text{te}})$, where $P(\mathcal{G}, \mathcal{Y})$ represents the distribution of the graphs and their labels. 
The objective of OOD generalization can be expressed as:
\begin{equation}
    \arg\min_F \mathbf{E}_{\mathcal{G}, \mathcal{Y} \sim P(\mathcal{G}_{\text{te}}, \mathcal{Y}_{\text{te}})} \left[l \left(F(\mathcal{G}), \mathcal{Y}\right) \mid \mathcal{G}_{\text{tr}}, \mathcal{Y}_{\text{tr}}\right],
    \label{eq:ood}
\end{equation}
where $l: \mathcal{Y} \times \mathcal{Y} \to \mathbb{R}$ is a loss function.
In this paper, we explore a setting where neither the test graphs $\mathcal{G}_{\text{te}}$ nor their corresponding labels $\mathcal{Y}_{\text{te}}$ are available during the training phase. 

\subsection{Differentiable GraphNAS}
Unlike traditional GraphNAS approaches, which treat selecting the best architecture as a black-box optimization problem within a discrete domain, differentiable GraphNAS\cite{qin2022graph,yao2024data} relaxes the discrete search space into a continuous one and allows for efficient optimization via gradient descent. We define the set of candidate operations as $\mathcal{O} = \{o_1, o_2, \ldots, o_K\}$, where each $o_k \in \mathcal{O}$ represents an operation from the search space, and $K$ is the total number of operations in $\mathcal{O}$. 
Furthermore, differentiable GraphNAS relaxes the rigid selection of operations in $\mathcal{O}$ into a soft selection where each candidate is assigned a probability.
Eq.~\eqref{eq:diffGraphNAS} illustrates an example of differentiable search along the architecture space. The output of $l^{th}$ layer can be represented as:
\begin{equation}
    \mathbf{h}_i^{(l+1)} = \sum_{o \in \mathcal{O}} p^o  o(\mathbf{h}_i^{(l)}),
    \label{eq:diffGraphNAS}
\end{equation}
where $\mathbf{h}_i^{(l)}$ represents the embedding of node $i$ at the $l^{\text{th}}$ layer, and $p^{o}$ is the probability associated with the corresponding candidate operation $o$. The probability distribution across this dimension is normalized such that ${\sum}_{o \in \mathcal{O}} p^{o} = 1$. 
In each graph, all nodes share the same set of probabilities, and final architectures can be obtained by retaining the candidate operations with the highest probabilities during the testing process.

\subsection{Power Law Distributions and Assortativity}



\noindent \textbf{Power Law Distribution}
In many real-world graphs, such as social networks and molecular networks, the degree distribution of nodes often follows a power law.
This distribution indicates that the probability $P(k)$ of a node having $k$ connections is proportional to $k^{-\alpha}$, where $\alpha$ is a positive constant:
\begin{equation}
    P(k) \propto k^{-\alpha}
\end{equation}
The power-law distribution reflects the heterogeneity of node connectivity, where a small number of nodes (i.e., hubs) account for the majority of edges, while most nodes exhibit a long-tail degree distribution, as illustrated in Fig. \ref{fig:intro_draw}. To maximize the performance of GNNs, different message aggregation mechanisms can be employed for nodes at various positions within the power-law distribution. 

However, differences in global topological characteristics necessitate a differentiated treatment of the power-law distribution across different types of graphs.
Incorporating a global perspective enables the model to discern spurious motifs between the current graph structure and degree distribution.

\noindent \textbf{Assortativity} is a measure of the tendency of nodes in a graph to connect with other nodes that are similar in some specified attribute. This metric often reveals important structural patterns, such as the tendency of individuals in social networks to associate with others who are similar to themselves. Formally, the assortativity coefficient can be defined as:
\begin{equation}
    q_j = \frac{j p_j}{\sum_k k p_k},
    \label{eq:Ass1}
\end{equation}
\begin{equation}
    \gamma = \frac{\sum_{jk} jk \left(e_{jk} - q_j q_k\right)}{\sigma_q^2},
    \label{eq:Ass2}
\end{equation}
where $p_j$ is the proportion of nodes of degree $j$, $e_{jk}$ is the proportion of edges in the graph that connect nodes of degree $j$ to nodes of degree $k$, and $\sigma_q$ is the standard deviation of $q$. $\gamma$ typically ranges from $-1$ to $1$, where a positive $\gamma$ indicates that high-degree nodes tend to connect with other high-degree nodes, while a negative $\gamma$ suggests they connect with low-degree nodes. 
We leverage $\gamma$ to enable NodeNAS to learn invariant representations of graphs with distribution shifts, thus enabling the architectures searched with generalizability.

\section{Node-Specific Graph Neural Architecture Search}
\label{sec:nodenas}
NodeNAS introduces a paradigm shift from traditional GNN architectures, where a singular embedding update method is uniformly applied all nodes in the graph within each layer. 
In contrast, NodeNAS proposes a flexible and adaptive framework that dynamically tailors the appropriate node information aggregation method based on the specific needs of each node within the given graph.
Specifically, given a graph $\mathcal{G}=\{V,E\}$ with $V$ denoting the set of nodes and $E$ denoting the set of edges, NodeNAS aims to learn an architecture mapping function $\Phi: \mathcal{G} \rightarrow \mathcal{A} \times W_\mathcal{A}$, where $\mathcal{A}$ represents the architecture for each node i.e., $\mathcal{A}=\{ A_1,A_2,...,A_N \}$, and $W_\mathcal{A}$ the associated weights. $N=|V|$ denotes the number of nodes in $\mathcal{G}$. 

To ensure the differentiability of NodeNAS, we also introduce weight sharing and assign each node a probability vector $\mathbf{p}$. $\mathbf{p}$ represents the probabilities of different operations, such as $GATConv$, being applied to the node. 
Define the set of candidate operations as $\mathcal{O} = \{ o_1, o_2, \ldots, o_K \}$, where each $o_k$ in $\mathcal{O}$ represents an operation in the search space, and $K$ is the total number of operations. $A_i$ can be further express as ${A}_i = \{(o, p_i^{o})\}$, where $o$ is the candidate operations satisfying $o\in \mathcal{O}$ and $p_i^{o}$ is corresponding probability. In $l^{th}$ layer,  the embedding update for the node $i$ can be expressed as:
\begin{equation}
    \mathbf{h}_i^{(l+1)} = \sum_{o \in \mathcal{O}} p_i^o  o({\mathbf{h}}_i^{(l)}),
\end{equation}
where $\mathbf{h}_i^{(l)}$ represents the embedding of node $i$ at the $l^{th}$ layer. 
The weights of operation $o$ are shared across different graphs, ensuring that node-specific architectures can be tailored end-to-end for each graph, while enabling efficient optimization through gradient descent.

\section{Adaptive Aggregation Attention Based Multi-Dim NodeNAS}

\subsection{Framework}
In our proposed method, we tailor an unique node-specific architecture for each graph by maximizing the learning of intrinsic relationships between node and graph distributions from limited datasets.
In particular, our method supports searching the architectures across multiple dimensions, allowing for more flexibility in expressing learned decoupled information in architecture customization.

Specifically, we aim to learn an architecture mapping function ${\Phi}: G \rightarrow A\times\mathcal{W}_A$. Unlike Section \ref{sec:nodenas}, $A_i$ of node $i$ contains a set of architectures composed of multiple search dimensions, i.e., $A_i = \{ A_i^1, A_i^2, \ldots, A_i^Z \}$, where $Z$ is the number of S-Dims and $A_i^z = \{(o_k, p^{z,o_k})\}$ represents the architecture searched in the $z^{th}$ S-Dim.
$|A_i^z| = |\mathbf{p}| = K$ always holds for any $z$, indicating that the search space within each S-Dim includes the full set of candidate operations, thereby enabling the model to learn architectural preferences differentiably for different nodes across graphs.
$\Phi$ can further be decomposed into a set of mappings for each S-Dim, i.e., $\Phi=\{\Phi_1,\Phi_2,\ldots, \Phi_Z\}$, with each $\Phi_z$ mapping the graph to a suitable architecture $A_z$ and corresponding weights $\mathcal{W}_{A_z}$ in $z^{th}$ S-Dim.
Therefore, Eq.\ref{eq:ood} can be transformed into the following form:

\begin{equation}
    \underset{\Phi\subset{S_p}}{\rm min} \sum_{(g_i\,y_i) \in \mathcal{G}_{\text{tr}} } [ \mathcal{L}( F(\sum_{z=1}^Z {\Phi_z}(g_i),g_i),y_i) +\beta\mathcal{L}_{\rm reg}(\Phi(g_i)],
    \label{FG}
\end{equation}
where $\beta$ is a hyperparameter and $\mathcal{L}_{\text{reg}}\left(\Phi(g_i)\right)$ is the regularizer for the architectures. 
$S_p$ represents the search space, a two-dimensional space composed of the operation plane and the probability plane, and $F(\cdot)$ is used to obtain the output for downstream tasks under the tailored architectures and weights.

For each $g_j\in G_{\text{te}}$, we utilize the trained $\Phi$ to generate unique architectures that are tailored to $g_j$ and perform well under distribution shifts.
Especially, for certain unsupervised tasks, such as community detection and graph partitioning, $\mathcal{G}_{\text{tr}}$ contains only a single graph and $y_i$ is not required.


\subsection{Disentangled Mapping Encoder}

\label{sec:encoder}
\begin{figure}[t]
	\centering
	\includegraphics[width=0.99\linewidth]{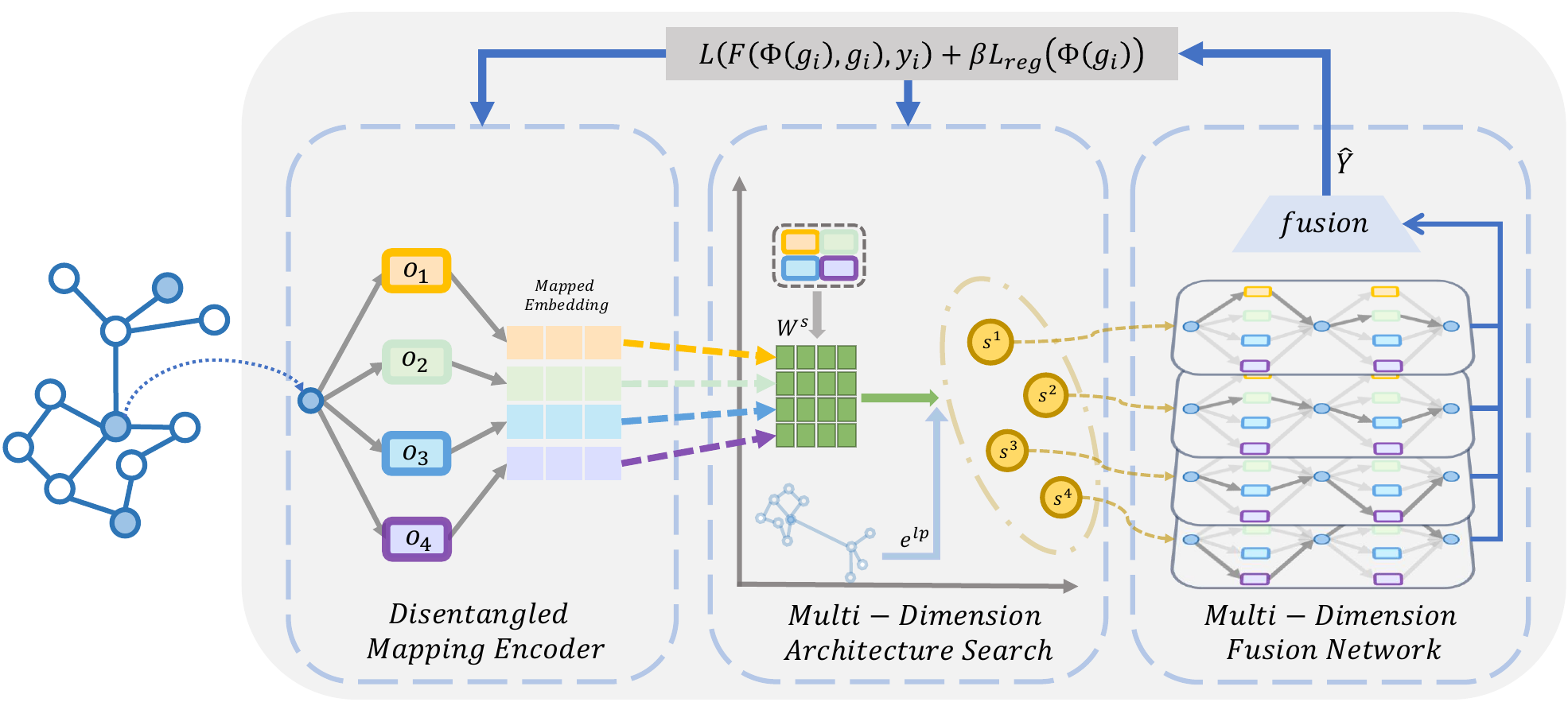}
	\caption{An overview of our proposed MNNAS model.}
 	\label{MNNAS}
\end{figure}
We aims for the encoder to rapidly aggregate overall semantic features. However, in GNNs, information is primarily propagated through edges, necessitating a deep convolutional network to capture global information. 
Moreover, the multi-layer fixed GNN architecture exacerbates the nonlinear dependency between the input features and the adjacency matrix, which arises due to spurious motifs in the distribution.
Therefore, our encoder only takes features as input and manually aggregates the global semantic information to accelerate information propagation between nodes. Specifically, our encoder is designed as follows:

\begin{equation}
    \mathbf{h}_i^{(l)} = \text{ENCODER}(\mathbf{h}_i^{(l-1)} +  \eta \ {\text{LINEAR}}(\frac{1}{N}\sum_{i=1}^{N} \mathbf{h}_i^{(l-1)})),
    \label{eq:gnn_update}
\end{equation}
where $\eta$ is a initial hyperparameter and $\mathbf{h}_i^{(l)}$ is the embedding of node $i$ at $l^{th}$ layer. $\mathbf{h}_i^{(0)}$ is initialized with the features $\mathbf{x}_i$ of node $i$.

After obtaining the node embeddings, we map candidate operations to learnable embeddings for each node. Specifically, for each operation $o$ in $\mathcal{O}$, we learn a corresponding embedding $\mathbf{e}_i^o$.
During the architecture search, we replace $\mathcal{O}$ with the set $E_i$ which is composed of mapped embeddings.
The mapping from operations to embeddings denoted by $\Phi_{\mathcal{O}\rightarrow E}$ can be expressed as:
\begin{equation}
        E_i=[\mathbf{e}_i^1,...,\mathbf{e}_i^K]=[ o_1(\mathbf{h}_i^{(l)},...,o_K(\mathbf{h}_i^{(l)}) ],
        \label{eq:mapping_embeddings}
\end{equation}
where $\mathbf{e}_i^{k}$ represents the mapped embedding of $o_k$ for node $i$ in $l^{th}$ layer and $k$ is the number of candidate operations. $l$ is omitted here to simplify the expression. 
$\Phi_{\mathcal{O}\rightarrow E}$ allows the model to perform multi-dimension search under single-dimension computation complexity.
This is facilitated by that the mapping is shared among different S-Dims, i.e. $\mathbf{e}_{i(dim_a)}^{k} = \mathbf{e}_{i(dim_b)}^{k} = \mathbf{e}_i^k$ always holds in both the $a^{th}$ S-Dim and $b^{th}$ S-Dim. Each operation can simultaneously participate in the architecture search of multiple S-Dims but is computed only once. 

Furthermore, to prevent mode collapse, where the mapped embeddings of different operations trend to be indistinguishable during training, we incorporate a regularization term that leverages cosine distance to maintain diversity among mapped embeddings:
\begin{equation}
    L_{\text{cos}} = \sum_i \sum_{\substack{o, o' \in \mathcal{O} \\ o \neq o'}} \frac{\mathbf{e}_i^o \cdot \mathbf{e}_i^{o'}}{\|\mathbf{e}_i^o\|_2 \|\mathbf{e}_i^{o'}\|_2},
    \label{eq:cos_loss}
\end{equation}
where $\mathbf{e}_i^o$ and $\mathbf{e}_i^{o'}$ denote the embeddings of node $i$ for operations $o$ and $o'$, respectively. $L_{\text{cos}}$ ensures orthogonality of $\Phi_{\mathcal{O}\rightarrow E}$, facilitating targeted encoding of disentangled information within the graph.

\subsection{Multi-Dimension Architecture Search}
\label{sec:aggregation}

We propose a multi-dimension architecture customization method, mapping each graph to node-specific GNN architectures across multiple S-Dims, i.e., tailoring multiple sets of $A_i^k$. 
It includes two components: the link pattern encoder, which learns structure-invariant information across different graph distributions that follow power-law distribution, and adaptive aggregation attention, which guides the model to search architecture across multiple S-Dims simultaneously.

\noindent \textbf{Link Pattern Encoder} 
With limited datasets, node-specific architecture search could better express the disentangled information, while the multi-dim architecture amplifies the generalization brought by NodeNAS. 
However, capturing such disentangled information hidden in the distribution is challenging. 
When addressing spurious correlations in the training set, it is crucial to distinguish the preferences of nodes with varying degrees within similar graphs and those of nodes with similar degrees across different graphs.

To address these issues, we incorporate the degree distribution of different nodes into the link pattern encoder and use assortativity to quantify the overall graph structure, promoting the disentanglement of nodes and graphs in terms of topology. Specially, the link pattern encoder can be express as:
\begin{equation}
    \mathbf{e}_i^{lp} = \text{ENCODER}(\gamma_{g}, \ \frac{d_i^2}{\bar{d}^2}, \ \frac{d_i}{\bar{d}},\  \frac{1}{|E|}\sum_{(a,b) \in E} d_a d_b),\ i\in g,
\end{equation}
where $d_a$ and $d_b$ are the degrees of nodes $a$ and $b$, respectively, connected by an edge. $\bar{d}$ and $\bar{d}^2$ are the mean degree and mean square degree, respectively, of all nodes in graph $g$ where node $i$ resides.
Especially, we approximate assortativity coefficient as a function of degree statistics, i.e., rewriting Eq. \ref{eq:Ass2} as follow:
\begin{equation}
\gamma_g \approx \frac{\frac{1}{|E|} \sum_{(a,b) \in E} d_a d_b - \left[\frac{1}{|E|} \sum_{(a,b) \in E} \frac{1}{2} (d_a + d_b)\right]^2}{\frac{1}{|E|} \sum_{(a,b) \in E} \frac{1}{2} (d_a^2 + d_b^2) - \left[\frac{1}{|E|} \sum_{(a,b) \in E} \frac{1}{2} (d_a + d_b)\right]^2},
\label{eq:assortativity_estimation}
\end{equation}
where the numerator denotes the actual versus expected differences in degrees of connected node and the denominator denotes the statistical properties of the degree distribution.

\noindent \textbf{Adaptive Aggregation Attention} 
Adaptive aggregation attention combines link pattern vectors $ \mathbf{e}^{lp} $, searching probability values for each candidate operation in every S-Dim. Specifically, for node $ i $ in the $ z^{th}$ S-Dim, the probability of different operations in $\mathcal{O}$ can be calculated as follow:

\begin{equation}
    \mathbf{s}_{i}^{z}=\mathbf{e}_i^{lp} \odot 
    \frac{\mathbf{e}_i^z W^s (E_i) ^{\top}}{\sqrt d}, \
    {p}_{i}^{z,o}=\frac{\text{exp}(s_{i}^{z,o})}{\sum_{o^\prime \in \mathcal{O}} \text{exp}(s^{z,o^\prime}_i)},
    \label{attention}
\end{equation}
where $\odot$ denotes Hadamard Product and $\mathbf{s}_i^z$ denotes the operation preference vector representing the architecture subspace, which is the projection of candidate operations onto the $z^{th}$ S-Dim.
$p_{i}^{z,o}$ represents the probability of operation $o$ in $z^{th}$ S-Dim for node $i$. 
$d=|\mathbf{e}_i^z|$ is the dimension of the mapped embedding in Eq. \ref{eq:mapping_embeddings}, and $W^s$ is the weight matrix satisfying $W^s \in \mathbb{R}^{d \times d}$.
Eq. \ref{attention} imposes a constraint that the number of dimensions we search must always equal the number of candidate operations in $\mathcal{O}$, denoted as $|S_i|=Z=|\mathcal{O}|=K$. The attention indirectly provides a normalization constraint, preventing optimization dilemmas that could arise from indiscriminately expanding search dimensions in the context of limited datasets. 

Adaptive aggreation attention leverages the link pattern encoder to decorrelate node features from specific graph distributions and further quantify the importance of each mapped embedding within $E_i$. 
Essentially, this mechanism is based on $att\gets q \times k$, centered on different mapped embeddings to maximize the probability of similar representations learned through various operations.




\subsection{Multi-Dimension Fusion Network}
\label{sec:search}
We integrate architectures searched from different S-Dims into a continuous space, jointly considering architectures in all S-Dims to learn the final node representations.  The node representations learned after the fusion of multi-dimension architectures can be calculated as follows:
\begin{equation}
{\mathbf{h}}_i^{(l+1)} = \sigma(\frac{1}{Z} \sum_{z=1}^Z \sum_{o \in \mathcal{O}} p_i^{z,o}o(\mathbf{h}_i^{(l)}))+\mathbf{h}_i^{(l)},
\label{fuse}
\end{equation}
where $\mathbf{h}_i^{(l)}$ denotes the initial node embedding at the $l^{th}$ layer, and ${\mathbf{h}}_i^{(l+1)}$ represents the output node representation. $\sigma$ here denotes activation function. 
Due to the mapping function $\Phi_{\mathcal{O}\rightarrow E}$, we assign probabilities to mapped embeddings in $S_p$ within each S-Dim rather than to the operations, avoiding redundant computations and unnecessary learnable parameters.
Shortcut connections is utilized in Eq. \ref{fuse} to prevent overfitting and mitigate the influence of irrelevant features.

Our approach eliminates the retraining and architecture discretization steps common in NAS methods by maintaining continuous architectures with weight-sharing across graphs. This enables end-to-end execution through shared network parameters, effectively creating an ensemble model that bypasses separate architecture-specific training.

\subsection{Theoretical Insights}
\label{sec:generalization}
In this section, we present a theoretical analysis of how adaptive aggregation attention enhances the model's generalization capability through the lens of information bottleneck theory. 

Let us consider a node $i$ with an operation mapping set  $E_i$ and the corresponding output representation as $Z_i$. Thus, the mapping function can be expressed as  $f(Z_i|E_i)$ . We assume a distribution $E_i \sim \text{Gaussian}(E_i^{'},\epsilon)$, where $E_i$ represents the noisy input variable, $E_i^{'}$ is the invariant target variable, and $\epsilon$ denotes the variance of the Gaussian noise. The information bottleneck can be formulated as follows:
\begin{equation}
    f_{IB}(Z_i|E_i) = \arg\min_{f(Z_i|E_i)} I(E_i,Z_i) - I(Z_i,E_i^{'}),
\label{eq:IB}
\end{equation}
where \( I(\cdot,\cdot) \) denotes the mutual information. 
In the Eq.\ref{attention}, we decompose $W^s=W^q(W^o)^{\top}$ and transform the probability of operation $o$ to $(\mathbf{e}_i^z W^q) (\mathbf{e}_i^o ({e}_{i,o}^{lp} W^o))^{\top} $, and we get $S_i = Q(E_i)K^{\top}(E_i,\mathbf{e}_i^{lp})$.
Based on the derivation in \cite{zhou2022understanding,guo2024investigating}, we derive the iterative process for optimizing Eq.\ref{eq:IB} and Eq.\ref{attention}.
\begin{equation}
    Z_i = \sum_{o \in O} \frac{\exp(\mathbf{e}_i^z W^q) (\mathbf{e}_i^o ({e}_{i,o}^{lp} W^o))^{\top})}{\sum_{o' \in O} \exp(\mathbf{e}_i^z W^q) (\mathbf{e}_i^{o'} ({e}_{i,o'}^{lp} W^o))^{\top})} \mathbf{e}_i^o = \sum_{o \in O}
    p_i^{z,o}\mathbf{e}_i^o,
\label{eq:attention}
\end{equation}
where Eq. \ref{eq:attention} elucidates the relationship between attention mechanisms across operations and the information bottleneck. Furthermore, previous studies \cite{yang2023individual} has demonstrated the effectiveness of the information bottleneck for generalization, particularly in aiding GNNs to discard spurious features. Therefore, we assert that MNNAS facilitates generalization, a claim that is subsequently substantiated by extensive empirical evidence.

\subsection{Complexity Analysis}
Let $|V|$ and $|E|$ denote the number of nodes and edges in the graph, respectively. We define $d_i$ as the dimensionality of the input features, $d_e$ as the dimensionality of the initial node embeddings, $d_m$ as the dimensionality of the mapped embeddings for the different operations in $\mathcal{O}$, and $d_o$ as the dimensionality of the output.

\noindent \textbf{Number of Learnable Parameters:} In our framework, the node encoder module comprises $O(2 d_i d_e)$ parameters, the module for learning mapped embeddings includes $O(|\mathcal{O}|d_e d_m)$ learnable parameters, the link pattern encoder contains $4|\mathcal{O}|$ parameters, the adaptive aggregation attention has $O(d_m^2)$ parameters, and the multi-dimension fusion network has no learnable parameters. The final output layer consists of $O(d_m d_o)$ parameters. For an $\eta$-layer network, the total number of learnable parameters is given by:
$O(2 d_i d_e + d_m d_o + \eta (d_m^2 + |\mathcal{O}|(4 + d_e d_m))).$

\noindent \textbf{Time Complexity:} The time complexity of the node encoder is $O(|V|d_i d_e)$. For the mapped embeddings module, it is $O(|V||\mathcal{O}|d_e d_m)$. The multi-dimension architecture search module has a time complexity of $O(|V||\mathcal{O}|(d_m^2 + |\mathcal{O}|^2))$, and the multi-dimension fusion module's time complexity is $O(|V||\mathcal{O}|^2 d_m)$. The output layer has a time complexity of $O(|V|d_m d_o)$. Additionally, the time complexity for $L_{cos}$ is $O(|V||\mathcal{O}|^2 d_e)$.

Given that different S-Dims share operation weights through $\Phi$, and considering that $|\mathcal{O}|$ is a small constant, the time complexity  for multi-dimension search remain equivalent to those for single-dimension search, thus not incurring additional training costs. Consequently, the total time complexity can be simplified to:
$O(|V|(d_i d_e + |\mathcal{O}|d_e d_m + |\mathcal{O}|d_m^2 + d_m d_o)),$

\section{Experiments}

In this section, we report experimental results to verify the effectiveness of our model.

\noindent \textbf{Dataset:}
We established both synthetic and real-world datasets to evaluate the performance of MNNAS in supervised graph classification as well as unsupervised community detection and inverse graph partitioning tasks, specifically under conditions involving distribution shifts.

For graph classification, we use the Spurious-Motif (Motif) synthetic dataset, which integrates base and motif shapes, and the OGBG-Mol datasets (hiv, bace, sider) for molecular property predictions. Community detection utilizes the Cora, CiteSeer, and PubMed datasets. For inverse graph partitioning tasks, synthetic datasets including Erdős-Rényi (ER), random regular (RR), Barabási-Albert (BA), and Newman-Watts-Strogatz (NW) graphs are employed. 

\begin{wrapfigure}{r}{0.5\textwidth} 
  \centering
  \includegraphics[width=0.48\textwidth]{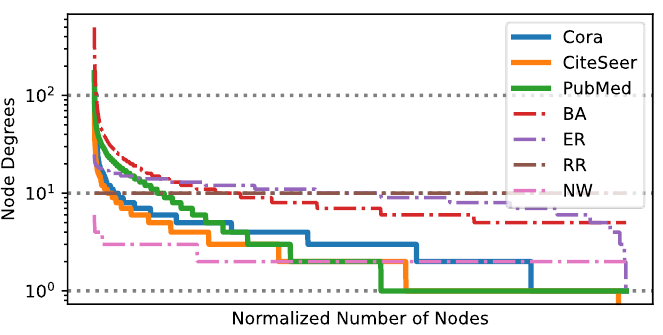} 
  \caption{Degree distribution of datasets.}
  \label{fig:deg-dis}
\end{wrapfigure}


\noindent \textbf{Baselines:}
We compare our model with the following baselines.
\begin{enumerate}
\item \textbf{Outstanding GNNs:} Commonly used manually designed GNNs include GCN\cite{kipf2016semi}, GAT\cite{velivckovic2017graph}, GIN\cite{xu2018powerful}, GraphSAGE\cite{hamilton2017inductive}, GraphConv\cite{morris2019weisfeiler} , GAP \cite{nazi1903gap} and ClusterNet\cite{wilder2019end}. Recently proposed methods like ASAP\cite{ranjan2020asap}, DIR\cite{wu2022discovering}, PNA\cite{corso2020principal}, and GSAT\cite{miao2022interpretable} have demonstrated strong performance in graph-level tasks with OOD settings.
\item \textbf{Outstanding NAS:} Our study evaluates six advanced NAS baselines, including DARTS\cite{liu2018darts} and five GraphNAS based algorithms: GraphNAS\cite{gao2021graph}, PAS\cite{wei2021pooling}, GASSO\cite{qin2021graph}, GRACES\cite{qin2022graph}, and DCGAS\cite{yao2024data}, which is currently the state-of-the-art in GNAS.
\end{enumerate}

\begin{table}[t]
    \centering
    \scriptsize
    \caption{Dataset Statistics.}
    \label{tab:datasets}
    \begin{tabularx}{\textwidth}{@{}l *{11}{>{\centering\arraybackslash}X}@{}} 
        \toprule
        \textbf{Dataset} & \textbf{Motif} & \textbf{hiv} & \textbf{bace} & \textbf{sider} & \textbf{Cora} & \textbf{Cite} & \textbf{Pub} & \textbf{BA} & \textbf{ER} & \textbf{RR} & \textbf{NW} \\ 
        \midrule
        Setting & Sup. & Sup. & Sup. & Sup. & Unsup. & Unsup. & Unsup. & Unsup. & Unsup. & Unsup. & Unsup. \\ 
        Task & Graph & Graph & Graph & Graph & Node & Node & Node & Node & Node & Node & Node \\ 
        Graphs & 18,000 & 41,127 & 1,513 & 1,427 & 1 & 1 & 1 & 1 & 1 & 1 & 1 \\ 
        Avg. Nodes & 26.1 & 25.5 & 34.1 & 33.6 & 2,708 & 3,327 & 19,717 & 10,000 & 10,000 & 10,000 & 10,000 \\ 
        Avg. Edges & 36.3 & 27.5 & 36.9 & 35.4 & 10,556 & 9,104 & 88,648 & 99,950 & 99,950 & 100,000 & 22,092 \\ 
        Classes & 3 & 2 & 2 & 2 & 7 & 6 & 3 & - & - & - & - \\ 
        \bottomrule
    \end{tabularx}
\end{table}

\begin{table}[t]
\begin{center}
\caption{Test Accuracy on Spurious-Motif and Test AUC-ROC on OGBG-Mol.}
\label{tab:gc}
\begin{tabularx}{\textwidth}{@{}>{\centering\arraybackslash}l|>{\centering\arraybackslash}X>{\centering\arraybackslash}X>{\centering\arraybackslash}X||>{\centering\arraybackslash}X>{\centering\arraybackslash}X>{\centering\arraybackslash}X@{}}
\toprule
\multirow{2}{*}{\textbf{Method}} & \multicolumn{3}{c||}{\textbf{Spurious-Motif (Accuracy)}} & \multicolumn{3}{c}{\textbf{OGBG-Mol (AUC-ROC)}} \\
                        & b=0.7       & b=0.8        & b=0.9      & hiv & sider & bace \\ \midrule
GCN                       & $48.39_{\pm 1.69}$           & $41.55_{\pm 3.88}$          & $39.13_{\pm 1.76}$          & $75.99_{\pm 1.19}$ & $59.84_{\pm 1.54}$ & $68.93_{\pm 6.95}$ \\
GAT                       & $50.75_{\pm 4.89}$          & $42.48_{\pm 2.46}$          & $40.10_{\pm 5.19}$          & $76.80_{\pm 0.58}$ & $57.40_{\pm 2.01}$ & $75.34_{\pm 2.36}$ \\
GIN                       & $36.83_{\pm 5.49}$          & $34.83_{\pm 3.10}$           & $37.45_{\pm 3.59}$          & $77.07_{\pm 1.49}$ & $57.57_{\pm 1.56}$ & $73.46_{\pm 5.24}$ \\
SAGE                    & $46.66_{\pm 2.51}$          & $44.50_{\pm 5.79}$           & $44.79_{\pm 4.83}$          & $75.58_{\pm 1.40}$ & $56.36_{\pm 1.32}$ & $74.85_{\pm 2.74}$ \\
GraphConv         & $47.29_{\pm 1.95}$          & $44.67_{\pm 5.88}$           & $44.82_{\pm 4.84}$          & $74.46_{\pm 0.86}$ & $56.09_{\pm 1.06}$ & $78.87_{\pm 1.74}$ \\
ASAP                    & $54.07_{\pm 13.85}$          & $48.32_{\pm 12.72}$          & $43.52_{\pm 8.41}$          & $73.81_{\pm 1.17}$ & $55.77_{\pm 1.18}$ & $71.55_{\pm 2.74}$ \\
DIR                       & $50.08_{\pm 3.46}$           & $48.22_{\pm 6.27}$           & $43.11_{\pm 5.43}$          & $77.05_{\pm 0.57}$ & $57.34_{\pm 0.36}$ & $76.03_{\pm 2.20}$ \\
\midrule
DARTS                   & $50.63_{\pm 8.90}$           & $45.41_{\pm 7.71}$           & $44.44_{\pm 4.42}$          & $74.04_{\pm 1.75}$ & $60.64_{\pm 1.37}$ & $76.71_{\pm 1.83}$ \\
GraphNAS                    & $55.18_{\pm 18.62}$          & $51.64_{\pm 19.22}$          & $37.56_{\pm 5.43}$          & - & - & - \\
PAS                     & $52.15_{\pm 4.35}$           & $43.12_{\pm 5.95}$           & $39.84_{\pm 1.67}$          & $71.19_{\pm 2.28}$ & $59.31_{\pm 1.48}$ & $76.59_{\pm 1.87}$ \\
GRACES         & $65.72_{\pm 17.47}$          & $59.57_{\pm 17.37}$          & $50.94_{\pm 8.14}$          & $77.31_{\pm 1.00}$ & $61.85_{\pm 2.56}$ & $79.46_{\pm 3.04}$ \\
DCGAS         & $87.68_{\pm 6.12}$          & $75.45_{\pm 17.40}$          & $61.42_{\pm 16.26}$          & \bm{$78.04_{\pm 0.71}$} & $63.46_{\pm 1.42}$ & $81.31_{\pm 1.94}$ \\
\midrule
\textbf{MNNAS}         & \bm{$97.53_{\pm 3.65}$}          & \bm{$98.42_{\pm 1.65}$}          & \bm{$93.19_{\pm 6.17}$}       & $76.55_{\pm 3.04}$ & \bm{$65.46_{\pm 1.18}$} & \bm{$84.69_{\pm 3.67}$} \\ \bottomrule
\end{tabularx}
\end{center}
\end{table}

\begin{table}[t]
    \centering
    \caption{Test Modularity on the real-world datasets.}
    \label{tab:cd}
    \begin{tabularx}{\textwidth}{@{} 
    l | *{3}{>{\centering\arraybackslash}X} |
    *{3}{>{\centering\arraybackslash}X} |
    *{3}{>{\centering\arraybackslash}X} 
    @{}}
        \toprule
         $\textbf{Method}$ &  
         $\textbf{Cora}_{\textbf{tr}}$ & $\textbf{Cite}_{\textbf{te}}$ &  $\textbf{Pub}_{\textbf{te}}$ &  $\textbf{Cite}_{\textbf{tr}}$ &   $\textbf{Cora}_{\textbf{te}}$ &   $\textbf{Pub}_{\textbf{te}}$ &  $\textbf{Pub}_{\textbf{tr}}$ &   $\textbf{Cite}_{\textbf{te}}$ &  $\textbf{Cora}_{\textbf{te}}$ \\ 
\midrule
GCN        & $0.65_{\scalebox{0.45}{$\pm 0.02$}}$ & $0.56_{\scalebox{0.45}{$\pm 0.01$}}$ & $0.50_{\scalebox{0.45}{$\pm 0.02$}}$ & $0.65_{\scalebox{0.45}{$\pm 0.01$}}$ & $0.58_{\scalebox{0.45}{$\pm 0.02$}}$ & $0.50_{\scalebox{0.45}{$\pm 0.02$}}$ & $0.64_{\scalebox{0.45}{$\pm 0.01$}}$ & $0.55_{\scalebox{0.45}{$\pm 0.03$}}$ & $0.54_{\scalebox{0.45}{$\pm 0.02$}}$ \\ 
GAT        & $0.59_{\scalebox{0.45}{$\pm 0.03$}}$ & $0.52_{\scalebox{0.45}{$\pm 0.05$}}$ & $0.42_{\scalebox{0.45}{$\pm 0.04$}}$ & $0.61_{\scalebox{0.45}{$\pm 0.04$}}$ & $0.50_{\scalebox{0.45}{$\pm 0.05$}}$ & $0.48_{\scalebox{0.45}{$\pm 0.02$}}$ & $0.60_{\scalebox{0.45}{$\pm 0.02$}}$ & $0.52_{\scalebox{0.45}{$\pm 0.01$}}$ & $0.49_{\scalebox{0.45}{$\pm 0.04$}}$ \\
GIN        & $0.58_{\scalebox{0.45}{$\pm 0.08$}}$ & $0.52_{\scalebox{0.45}{$\pm 0.06$}}$ & $0.45_{\scalebox{0.45}{$\pm 0.03$}}$ & $0.66_{\scalebox{0.45}{$\pm 0.03$}}$ & $0.52_{\scalebox{0.45}{$\pm 0.03$}}$ & $0.41_{\scalebox{0.45}{$\pm 0.05$}}$ & $0.56_{\scalebox{0.45}{$\pm 0.08$}}$ & $0.53_{\scalebox{0.45}{$\pm 0.07$}}$ & $0.49_{\scalebox{0.45}{$\pm 0.08$}}$ \\
SAGE       & $0.60_{\scalebox{0.45}{$\pm 0.03$}}$ & $0.47_{\scalebox{0.45}{$\pm 0.02$}}$ & $0.44_{\scalebox{0.45}{$\pm 0.03$}}$ & $0.64_{\scalebox{0.45}{$\pm 0.03$}}$ & $0.51_{\scalebox{0.45}{$\pm 0.05$}}$ & $0.50_{\scalebox{0.45}{$\pm 0.03$}}$ & $0.60_{\scalebox{0.45}{$\pm 0.03$}}$ & $0.48_{\scalebox{0.45}{$\pm 0.04$}}$ & $0.48_{\scalebox{0.45}{$\pm 0.03$}}$ \\
GraphConv  & $0.57_{\scalebox{0.45}{$\pm 0.03$}}$ & $0.42_{\scalebox{0.45}{$\pm 0.03$}}$ & $0.41_{\scalebox{0.45}{$\pm 0.04$}}$ & $0.45_{\scalebox{0.45}{$\pm 0.22$}}$ & $0.30_{\scalebox{0.45}{$\pm 0.16$}}$ & $0.29_{\scalebox{0.45}{$\pm 0.15$}}$ & $0.53_{\scalebox{0.45}{$\pm 0.02$}}$ & $0.41_{\scalebox{0.45}{$\pm 0.03$}}$ & $0.39_{\scalebox{0.45}{$\pm 0.03$}}$ \\
MLP        & $0.64_{\scalebox{0.45}{$\pm 0.02$}}$ & $0.48_{\scalebox{0.45}{$\pm 0.03$}}$ & $0.28_{\scalebox{0.45}{$\pm 0.03$}}$ & $0.66_{\scalebox{0.45}{$\pm 0.04$}}$ & $0.47_{\scalebox{0.45}{$\pm 0.05$}}$ & $0.30_{\scalebox{0.45}{$\pm 0.06$}}$ & $0.58_{\scalebox{0.45}{$\pm 0.04$}}$ & $0.44_{\scalebox{0.45}{$\pm 0.05$}}$ & $0.34_{\scalebox{0.45}{$\pm 0.03$}}$ \\
ClusterNet & $0.59_{\scalebox{0.45}{$\pm 0.02$}}$ & $0.56_{\scalebox{0.45}{$\pm 0.06$}}$ & $0.42_{\scalebox{0.45}{$\pm 0.04$}}$ & $0.70_{\scalebox{0.45}{$\pm 0.03$}}$ & $0.46_{\scalebox{0.45}{$\pm 0.01$}}$ & $0.31_{\scalebox{0.45}{$\pm 0.06$}}$ & $0.61_{\scalebox{0.45}{$\pm 0.02$}}$ & $0.58_{\scalebox{0.45}{$\pm 0.06$}}$ & $0.47_{\scalebox{0.45}{$\pm 0.03$}}$ \\
\midrule
DARTS      & $0.66_{\scalebox{0.45}{$\pm 0.02$}}$ & $0.53_{\scalebox{0.45}{$\pm 0.03$}}$ & $0.42_{\scalebox{0.45}{$\pm 0.06$}}$ & $0.67_{\scalebox{0.45}{$\pm 0.02$}}$ & $0.53_{\scalebox{0.45}{$\pm 0.03$}}$ & $0.47_{\scalebox{0.45}{$\pm 0.02$}}$ & $0.62_{\scalebox{0.45}{$\pm 0.04$}}$ & $0.54_{\scalebox{0.45}{$\pm 0.02$}}$ & $0.51_{\scalebox{0.45}{$\pm 0.03$}}$ \\
GASSO      & $0.63_{\scalebox{0.45}{$\pm 0.03$}}$ & $0.58_{\scalebox{0.45}{$\pm 0.04$}}$ & $0.52_{\scalebox{0.45}{$\pm 0.03$}}$ & $0.68_{\scalebox{0.45}{$\pm 0.03$}}$ & $0.57_{\scalebox{0.45}{$\pm 0.04$}}$ & $\bm{0.53_{\scalebox{0.45}{$\pm 0.04$}}}$ & $0.66_{\scalebox{0.45}{$\pm 0.04$}}$ & $0.61_{\scalebox{0.45}{$\pm 0.03$}}$ & $0.59_{\scalebox{0.45}{$\pm 0.03$}}$ \\
\midrule
\textbf{MNNAS}      & $\bm{0.69_{\scalebox{0.45}{$\pm 0.02$}}}$ & $\bm{0.63_{\scalebox{0.45}{$\pm 0.02$}}}$ & $\bm{0.53_{\scalebox{0.45}{$\pm 0.01$}}}$ & $\bm{0.72_{\scalebox{0.45}{$\pm 0.02$}}}$ & $\bm{0.60_{\scalebox{0.45}{$\pm 0.01$}}}$ & $0.52_{\scalebox{0.45}{$\pm 0.03$}}$ & $\bm{0.68_{\scalebox{0.45}{$\pm 0.01$}}}$ & $\bm{0.61_{\scalebox{0.45}{$\pm 0.01$}}}$ & $\bm{0.61_{\scalebox{0.45}{$\pm 0.02$}}}$ \\
\bottomrule
\end{tabularx}
\end{table}

\subsection{Graph Classification on Synthetic and Real Datasets}
\textbf{Experimental Setting} 
For the Spurious-Motif dataset, we followed the experimental setup outlined in\cite{qin2022graph}. Additionally, we performed experiments on real dataset OGBG-Mol. Each experiment was replicated ten times with different random seeds, and results are presented as averages with standard deviations.

\noindent \textbf{Qualitative Results:} Table \ref{tab:gc} shows that our model outperforms baseline methods on both synthetic and real datasets. Traditional GNNs underperform on synthetic data due to spurious correlations and distribution shifts. In real datasets, GNN effectiveness varies with graph characteristics. While NAS methods slightly improve upon manual GNN designs, they struggle with distribution changes. Conversely, MNNAS effectively reduces spurious correlations in graph distributions, notably improving performance, especially on synthetic datasets.

\subsection{Community Detection}
\textbf{Experimental Setting} We evaluate community detection performance on real datasets. Each dataset is partitioned into 10 communities as per the methodology described in the\cite{wilder2019end}.
To standardize feature dimensions across datasets, we employ Non-negative Matrix Factorization instead of original node features. 

\noindent \textbf{Qualitative Results} Table \ref{tab:cd} illustrates a notable decline in generalization performance among manually designed GNN models during tests. These models often perform well on training datasets but fail to maintain efficacy on testing datasets. 
A similar trend is observed with differentiable NAS methods, indicating that applying a uniform GNN approach across an entire graph diminishes generalization due to varying node preferences.
Conversely, our model consistently excels in both training and testing phases. The superior performance is attributed primarily to its capacity for node-specific architecture searches, which allows for the effective separation of invariant features specific to nodes. 

\begin{table}[t]
    \centering
    \caption{Test Ratio of inter-subgraph to total edges on the synthetic datasets.}
    \label{tab:igb}
    \begin{tabularx}{\textwidth}{@{} l | *{5}{>{\centering\arraybackslash}X} @{}}
        \toprule
        \textbf{Method} & $\textbf{BA1000}_{\textbf{tr}}$ & $\textbf{BA10000}_{\textbf{te}}$ & $\textbf{RR10000}_{\textbf{te}}$ & $\textbf{ER10000}_{\textbf{te}}$ & $\textbf{NW10000}_{\textbf{te}}$  \\ 
        \midrule
        GCN & $0.95_{\pm 0.01}$ & $0.86_{\pm 0.10}$ & $0.87_{\pm 0.07}$ & $0.87_{\pm 0.06}$ & $0.81_{\pm 0.11}$  \\ 
        GAT & $0.95_{\pm 0.01}$ & $0.87_{\pm 0.05}$ & $0.75_{\pm 0.07}$ & $0.75_{\pm 0.07}$ & $0.72_{\pm 0.09}$  \\ 
        SAGE & $\bm{0.99_{\pm 0.00}}$ & $0.95_{\pm 0.01}$ & $0.92_{\pm 0.04}$ & $0.90_{\pm 0.05}$ & $0.80_{\pm 0.09}$  \\ 
        GIN & $\bm{0.99_{\pm 0.00}}$ & $0.92_{\pm 0.00}$ & $0.80_{\pm 0.05}$ & $0.81_{\pm 0.04}$ & $0.66_{\pm 0.09}$  \\ 
        GraphConv & $\bm{0.99_{\pm 0.00}}$ & $0.94_{\pm 0.01}$ & $0.87_{\pm 0.05}$ & $0.87_{\pm 0.04}$ & $0.81_{\pm 0.04}$  \\ 
        MLP & $0.98_{\pm 0.00}$ & $0.92_{\pm 0.01}$ & $0.91_{\pm 0.01}$ & $0.90_{\pm 0.01}$ & $0.76_{\pm 0.06}$  \\ 
        GAP & $0.96_{\pm 0.01}$ & $0.91_{\pm 0.02}$ & $0.90_{\pm 0.02}$ & $0.90_{\pm 0.02}$ & $0.81_{\pm 0.10}$  \\
        \midrule
        \textbf{MNNAS} & $\bm{0.99_{\pm 0.00}}$ & $\bm{0.96_{\pm 0.00}}$ & $\bm{0.98_{\pm 0.00}}$ & $\bm{0.97_{\pm 0.00}}$ & $\bm{0.84_{\pm 0.05}}$  \\ 
        \bottomrule
    \end{tabularx}
\end{table}


\subsection{Inverse Graph Partition}
\textbf{Experimental Setting} 
To assess model generalization beyond typical power-law distributed datasets, we use four synthetic graphs with diverse distributions: BA, ER, RR, and NW. Training is conducted on a BA graph with 1,000 nodes, while testing uses larger graphs of 10,000 nodes
. Each graph is split into 10 subgraphs to maximize inter-subgraph edges and minimize intra-subgraph edges. 

\noindent \textbf{Qualitative Results} 
The results in Table \ref{tab:igb} demonstrate that employing a single GNN for generalization across datasets with diverse node distributions significantly degrades performance. This is due to the requirement for models to adapt to varying graph distributions globally and to distinct structural node features locally. Our proposed MNNAS model addresses this challenge by incorporating a link pattern encoder that decouples invariant structural factors, allowing for tailored architecture customization at the node level.

\begin{figure}[t]
	\centering
	\includegraphics[width=0.99\linewidth]{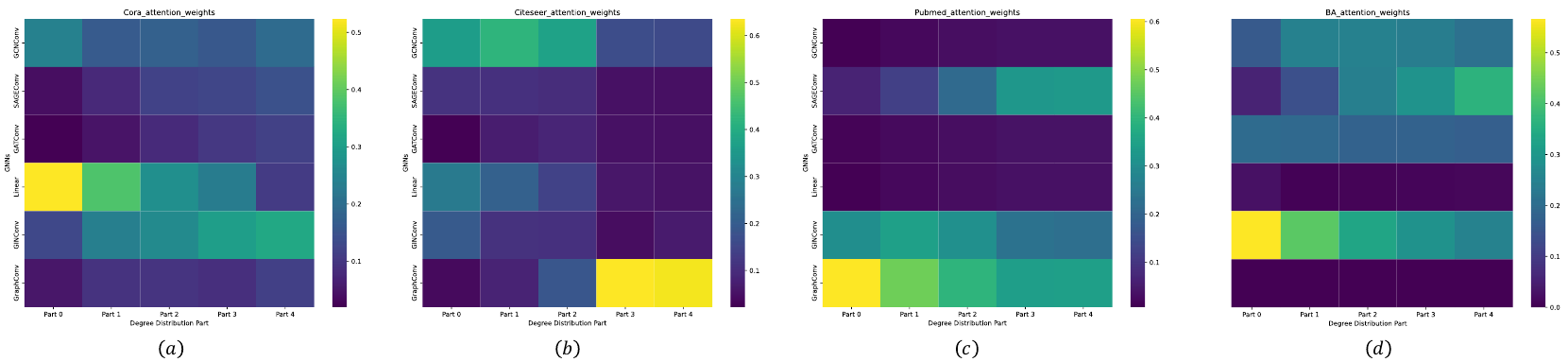}
	\caption{(a)-(d) illustrate the statistically derived architecture customization patterns of MNNAS for different degree distributions across datasets}
 	\label{fig:att}
\end{figure}

\subsection{Interpretability}
In Fig. \ref{fig:att}, we illustrates the architectural preferences of nodes across four datasets, categorized into five groups based on their degree, from highest to lowest. For Cora, high-degree nodes favor Linear architectures, while low-degree nodes prefer GINConv. In Citeseer, as the degree distribution shifts, architectural preferences transition from GCNConv to GraphConv; similar patterns are observed in Pubmed and BA.
These results indicate that nodes within each graph exhibit distinct architectural preferences influenced by their degree, underscoring the significance of power-law distributions. Moreover, nodes across different graphs demonstrate varying preferences, highlighting the effectiveness of incorporating graph-level topological features, such as assortativity, in differentiating power-law distributions across diverse graphs.

\section{Related Work}
\noindent\textbf{Graph Neural Architecture Search:}
Neural Architecture Search (NAS) automates creating optimal neural networks using RL-based\cite{zoph2018learning}, evolutionary\cite{miikkulainen2024evolving}, and gradient-based methods\cite{liu2018darts}.
Building on NAS's foundations,
GraphNAS\cite{gao2021graph} was the first to employ reinforcement learning for aggregating GNN architectures in the search space. AGNN\cite{zhou2022auto} introduced a Recurrent Neural Network controller to minimize noise in architecture search, alongside several notable works\cite{ding2020propagation,nunes2020neural}.
Additionally, PDNAS\cite{zhao2020probabilistic} pioneered differentiable search in GraphNAS, converting the discrete search space into a continuous one using Gumbel-Sigmoid. Furthermore, GRACES\cite{qin2022graph} is the first GraphNAS to address graph classification on OOD distributed datasets.

\noindent\textbf{GNN Generalization:}
Graph Neural Networks face representational dependencies that hinder their ability to generalize to unknown network structures, often resulting in poor performance on non-I.I.D. graphs\cite{hu2020open}. Recent research has aimed at improving GNN architectures for better performance under distribution shifts, including methods that integrate stochastic attention mechanisms\cite{miao2022interpretable}, random Fourier features\cite{xu2020neural}.
Additionally, DCGAS\cite{yao2024data} builds on GRACES by introducing a diffusion model-based data augmentation module, effectively improving classification accuracy on non-I.I.D. graphs.

\section{Conclusion}
In this paper, we present the Multi-dimension Node-specific graph Neural Architecture Search (MNNAS) framework, designed to enhance the generalization capabilities of GNAS in the face of distribution shifts. By customizing node-specific architectures that reflect the inherent variability of graph structures, MNNAS overcomes the limitations of existing GNAS methods, which typically depend on large training datasets and struggle with distribution patterns across different graphs. Our extensive experimental evaluations demonstrate that MNNAS, incorporating an adaptive aggregation attention mechanism and modeling power-law distributions, achieves superior performance across various supervised and unsupervised tasks under out-of-distribution conditions.


\begin{credits}
\subsubsection{\ackname} This work was partly supported by the National Natural Science Foundation of China under grants 72374154 and 62273260.
\end{credits}

%
%

\bibliographystyle{splncs04}
\bibliography{bibtex/main}

\begin{thebibliography}{10}
\providecommand{\url}[1]{\texttt{#1}}
\providecommand{\urlprefix}{URL }
\providecommand{\doi}[1]{https://doi.org/#1}

\bibitem{corso2020principal}
Corso, G., Cavalleri, L., Beaini, D., Li{\`o}, P., Veli{\v{c}}kovi{\'c}, P.: Principal neighbourhood aggregation for graph nets. Proc. of NeurIPS pp. 13260--13271 (2020)

\bibitem{ding2020propagation}
Ding, Y., Yao, Q., Zhang, T.: Propagation model search for graph neural networks. arXiv preprint arXiv:2010.03250  (2020)

\bibitem{gao2021graph}
Gao, Y., Yang, H., Zhang, P., Zhou, C., Hu, Y.: Graph neural architecture search. In: Proc. of IJCAI (2021)

\bibitem{guo2024investigating}
Guo, K., Wen, H., Jin, W., Guo, Y., Tang, J., Chang, Y.: Investigating out-of-distribution generalization of gnns: An architecture perspective. In: Proc. of KDD. pp. 932--943 (2024)

\bibitem{hamilton2017inductive}
Hamilton, W., Ying, Z., Leskovec, J.: Inductive representation learning on large graphs. Proc. of NeurIPS  (2017)

\bibitem{hu2020open}
Hu, W., Fey, M., Zitnik, M., Dong, Y., Ren, H., Liu, B., Catasta, M., Leskovec, J.: Open graph benchmark: Datasets for machine learning on graphs. Proc. of NeurIPS pp. 22118--22133 (2020)

\bibitem{kipf2016semi}
Kipf, T.N., Welling, M.: Semi-supervised classification with graph convolutional networks. arXiv preprint arXiv:1609.02907  (2016)

\bibitem{liu2018darts}
Liu, H., Simonyan, K., Yang, Y.: Darts: Differentiable architecture search. arXiv preprint arXiv:1806.09055  (2018)

\bibitem{miao2022interpretable}
Miao, S., Liu, M., Li, P.: Interpretable and generalizable graph learning via stochastic attention mechanism. In: Proc. of ICML. pp. 15524--15543 (2022)

\bibitem{miikkulainen2024evolving}
Miikkulainen, R., Liang, J., Meyerson, E., Rawal, A., Fink, D., Francon, O., Raju, B., Shahrzad, H., Navruzyan, A., Duffy, N., et~al.: Evolving deep neural networks. In: Artificial intelligence in the age of neural networks and brain computing, pp. 269--287 (2024)

\bibitem{morris2019weisfeiler}
Morris, C., Ritzert, M., Fey, M., Hamilton, W.L., Lenssen, J.E., Rattan, G., Grohe, M.: Weisfeiler and leman go neural: Higher-order graph neural networks. In: Proc. of AAAI. pp. 4602--4609 (2019)

\bibitem{nazi1903gap}
Nazi, A., Hang, W., Goldie, A., Ravi, S., Mirhoseini, A.: Gap: Generalizable approximate graph partitioning framework. arxiv 2019. arXiv preprint arXiv:1903.00614

\bibitem{nunes2020neural}
Nunes, M., Pappa, G.L.: Neural architecture search in graph neural networks. In: Intelligent Systems: 9th Brazilian Conference, BRACIS 2020, Rio Grande, Brazil, October 20--23, 2020, Proceedings, Part I 9. pp. 302--317 (2020)

\bibitem{qin2021graph}
Qin, Y., Wang, X., Zhang, Z., Zhu, W.: Graph differentiable architecture search with structure learning. Proc. of NeurIPS pp. 16860--16872 (2021)

\bibitem{qin2022graph}
Qin, Y., Wang, X., Zhang, Z., Xie, P., Zhu, W.: Graph neural architecture search under distribution shifts. In: Proc. of ICML. pp. 18083--18095 (2022)

\bibitem{ranjan2020asap}
Ranjan, E., Sanyal, S., Talukdar, P.: Asap: Adaptive structure aware pooling for learning hierarchical graph representations. In: Proc. of AAAI. pp. 5470--5477 (2020)

\bibitem{velivckovic2017graph}
Veli{\v{c}}kovi{\'c}, P., Cucurull, G., Casanova, A., Romero, A., Lio, P., Bengio, Y.: Graph attention networks. arXiv preprint arXiv:1710.10903  (2017)

\bibitem{wei2021pooling}
Wei, L., Zhao, H., Yao, Q., He, Z.: Pooling architecture search for graph classification. In: Proc. of CIKM. pp. 2091--2100 (2021)

\bibitem{wilder2019end}
Wilder, B., Ewing, E., Dilkina, B., Tambe, M.: End to end learning and optimization on graphs. Proc. of NeurIPS  (2019)

\bibitem{wu2022discovering}
Wu, Y.X., Wang, X., Zhang, A., He, X., seng Chua, T.: Discovering invariant rationales for graph neural networks. In: ICLR (2022)

\bibitem{xu2018powerful}
Xu, K., Hu, W., Leskovec, J., Jegelka, S.: How powerful are graph neural networks? arXiv preprint arXiv:1810.00826  (2018)

\bibitem{xu2020neural}
Xu, K., Zhang, M., Li, J., Du, S.S., Kawarabayashi, K.i., Jegelka, S.: How neural networks extrapolate: From feedforward to graph neural networks. arXiv preprint arXiv:2009.11848  (2020)

\bibitem{yang2023individual}
Yang, L., Zheng, J., Wang, H., Liu, Z., Huang, Z., Hong, S., Zhang, W., Cui, B.: Individual and structural graph information bottlenecks for out-of-distribution generalization. IEEE Transactions on Knowledge and Data Engineering  (2023)

\bibitem{yao2024data}
Yao, Y., Wang, X., Qin, Y., Zhang, Z., Zhu, W., Mei, H.: Data-augmented curriculum graph neural architecture search under distribution shifts  (2024)

\bibitem{zhao2020probabilistic}
Zhao, Y., Wang, D., Gao, X., Mullins, R., Lio, P., Jamnik, M.: Probabilistic dual network architecture search on graphs. arXiv preprint arXiv:2003.09676  (2020)

\bibitem{zhou2022understanding}
Zhou, D., Yu, Z., Xie, E., Xiao, C., Anandkumar, A., Feng, J., Alvarez, J.M.: Understanding the robustness in vision transformers. In: Proc. of ICML. pp. 27378--27394 (2022)

\bibitem{zhou2022auto}
Zhou, K., Huang, X., Song, Q., Chen, R., Hu, X.: Auto-gnn: Neural architecture search of graph neural networks. Frontiers in big Data p. 1029307 (2022)

\bibitem{zoph2018learning}
Zoph, B., Vasudevan, V., Shlens, J., Le, Q.V.: Learning transferable architectures for scalable image recognition. In: Proc. of CVPR. pp. 8697--8710 (2018)

\end{thebibliography}

\end{document}